\documentclass[conference]{IEEEtran}
\IEEEoverridecommandlockouts
\usepackage{times}
\usepackage{latexsym}
\usepackage{url}

\usepackage{hyperref}

\usepackage{makecell}
\usepackage{multirow}

\usepackage{graphicx}

\usepackage{enumitem}

\usepackage{booktabs}
\usepackage{siunitx}
\usepackage{etoolbox}
\robustify\bfseries

\usepackage{xspace}
\makeatletter
\DeclareRobustCommand\onedot{\futurelet\@let@token\@onedot}
\def\@onedot{\ifx\@let@token.\else.\null\fi\xspace}

\makeatother

\begin{document}

\title{Application of Low-resource Machine Translation Techniques to Russian-Tatar Language Pair}

\author{\IEEEauthorblockN{1\textsuperscript{st} Aidar Valeev}
\IEEEauthorblockA{\textit{Innopolis University}\\
Innopolis, Russia \\
ai.valeev@innopolis.ru}
\and
\IEEEauthorblockN{2\textsuperscript{nd} Ilshat Gibadullin}
\IEEEauthorblockA{\textit{Innopolis University}\\
Innopolis, Russia \\
i.gibadullin@innopolis.ru}
\and
\IEEEauthorblockN{3\textsuperscript{rd} Albina Khusainova}
\IEEEauthorblockA{\textit{Innopolis University}\\
Innopolis, Russia \\
a.khusainova@innopolis.ru}
\and
\IEEEauthorblockN{4\textsuperscript{th} Adil Khan}
\IEEEauthorblockA{\textit{Innopolis University}\\
Innopolis, Russia \\
a.khan@innopolis.ru}
}

\maketitle

\begin{abstract}
Neural machine translation is the current state-of-the-art in machine translation. Although it is successful in a resource-rich setting, its applicability for low-resource language pairs is still debatable. In this paper, we explore the effect of different techniques to improve machine translation quality when a parallel corpus is as small as 324 000 sentences, taking as an example previously unexplored Russian-Tatar language pair. We apply such techniques as transfer learning and semi-supervised learning to the base Transformer model, and empirically show that the resulting models improve Russian to Tatar and Tatar to Russian translation quality by +2.57 and +3.66 BLEU, respectively.
\end{abstract}

\section{Introduction}

The introduction of neural network models led to a game-changing shift in the area of machine translation (MT). Since this is a recent change, there's a lot of ongoing work: new architectures are being proposed, different techniques from statistical machine translation are being adapted, new ideas emerge. However, the performance of neural machine translation (NMT) degrades rapidly as the parallel corpus size gets small. There is a number of techniques to mitigate this problem, and we were interested in analyzing their effect for the Russian-Tatar language pair.


The Russian-Tatar language pair is virtually unexplored in machine translation literature, which is partially explained by the fact that the needed resources---both parallel and monolingual corpora are hard to find: first, there are few sources of such data, and second, they are mostly private. In general, there's a lack of research on translation between Slavic and Turkic languages, while it is an important direction for many countries in the post-Soviet area. In this paper, we present our results and analysis of applying different techniques to improve machine translation quality between Russian and Tatar languages given a small parallel corpus. Thus, our work may be especially useful for those who are interested in low-resource machine translation between Slavic and Turkic languages.

The paper is laid out as follows: Section \ref{section2} gives an overview of related work and languages' description; Section \ref{section3} describes the approaches we applied; Section \ref{section4} describes the data and the system; Section \ref{section5} reports the results and their analysis, and finally Section \ref{section6} gives some concluding remarks.

\section{\label{section2} Background and Related Work}
\subsection{NMT in Low-resource Setting}
Currently there are several fundamental neural machine translation models competing to be the state-of-the-art: recurrent \cite{bahdanau2015s2s} and convolutional \cite{gehring2017conv} neural networks, and more recent attention-based Transformer model \cite{vaswani2017transformer}. All models consist of encoder-decoder parts, but in the first one both encoder and decoder are RNN layers with an attention layer between them, the second one is based on convolutions, while Transformer uses stacks of self-attention and feed-forward layers. Transformer gives comparable results and takes less time to train, which is why we decided to use it as a base model.  

The performance of NMT models deteriorates sharply as the size of training data decreases, as it was observed by Koehn and Knowles \cite{koehn2017limit}. Different approaches were proposed to improve translation quality in a low-resource setting: Transfer Learning, Multitask Learning, Semi-supervised Learning, and Unsupervised Learning. Below we will quickly overview some of them.

One idea is to transfer knowledge learned from higher-resourced language pair. Zoph et al. \cite{zoph2016transfer} suggest to train a recurrent model \cite{bahdanau2015s2s} on a large parallel corpus of a rich language pair, then use it as initialization of weights for a low-resource language pair freezing the embeddings of the side, where language stays the same. Kocmi and Bojar \cite{kocmi2018trivial} apply the same idea but they don't freeze any parameters - they just change the training corpus. This helps when parent and child language pairs are not related or not aligned.

Zaremoodi et al. \cite{zaremoodi2018multitask} try to benefit from multitasking: instead of common recurrent unit of sequence-to-sequence model \cite{bahdanau2015s2s}, they use a unit with several blocks in it, one specific for translation task and others are shared between other tasks (Named-Entity Recognition, Syntactic or Semantic Parsing) using a routing network. All the tasks are trained together following the global objective, which is the weighted sum of the tasks' objectives.  

And, probably, the most popular idea is to exploit the power of monolingual data. Sennrich et al. \cite{sennrich2016bt} propose a semi-supervised method of using monolingual data, which is to pair monolingual sentences with their back-translation, mix the parallel corpus with the synthetic one and do not distinguish between them.

Next approach \cite{gulcehre} also exploits monolingual data, but in a different way: they train a language model (LM) on this data, where LM is a decoder part of the model without encoder input. They then integrate the LM with the translation model using Shallow Fusion - they get the final candidate words' scores by summing the NMT scores with the LM scores multiplied by a weighting hyper-parameter.

Stahlberg et al. \cite{stahlberg} apply the same idea but integrate language model differently (Simple Fusion with PostNorm): to calculate the final probability distribution they normalize the output of the projection layer of NMT using softmax to get the probability distribution, then multiply component-wise by the probability distribution of the LM predictions and normalize using softmax again. The work-flow is as follows: a LM is pre-trained on monolingual data separately, combined with a translation model, and finally, the combined model is trained on parallel data.

We use a combination of some of these ideas in our solution.

\subsection{Languages}
The Russian language belongs to the family of Indo-European languages, is one of the four living members of the East Slavic languages, and part of the larger Balto-Slavic branch, with 144 million speakers in Russia, Ukraine, and Belarus.

The Tatar language belongs to the family of Turkic languages, Common Turkic taxon, Kipchak sub-branch, Kipchak-Bulgar subdivision, with about 6 million speakers in and around the Tatarstan Republic, which is a part of the Russian Federation.

Both languages use the Cyrillic writing system, which is useful since the subwords vocabulary of the Transformer model is shared between source and target languages. Both languages are morphologically rich, and this results in a large number of word forms, complicating the translation. Also, Tatar is a gender-neutral language, while Russian is not, so, the translation could be biased.

For transfer learning, we use the related Kazakh language, which belongs to the family of Turkic languages, Common Turkic taxon, Kipchak sub-branch, Kipchak-Nogai subdivision, with about 10 million speakers in Kazakhstan and the Cyrillic writing system.

Both Tatar and Kazakh belong to the Kipchak group of Turkic languages. The spoken and written languages share some level of mutual intelligibility to native speakers, and alphabets significantly overlap.

\subsection{Russian-Tatar Machine Translation}
As we mentioned before, the Russian-Tatar language pair is virtually unexplored. To give an idea of the state-of-the-art: we found only one recent work on Russian-Tatar MT by Khusainov et al. \cite{khusainov}, where they augmented data by retranslating Turkic-Russian parallel corpora from Turkic to Tatar using rule-based systems. Also, there is a publicly available Russian-Tatar machine translation system by Yandex\footnote{\url{https://translate.yandex.ru}}, however, the quality is low for grammatically and syntactically complex sentences and this work is not described anywhere in the literature, so cannot be analyzed further.

\section{\label{section3} Methodology}
\begin{figure*}
\centering
\includegraphics[width=1\linewidth]{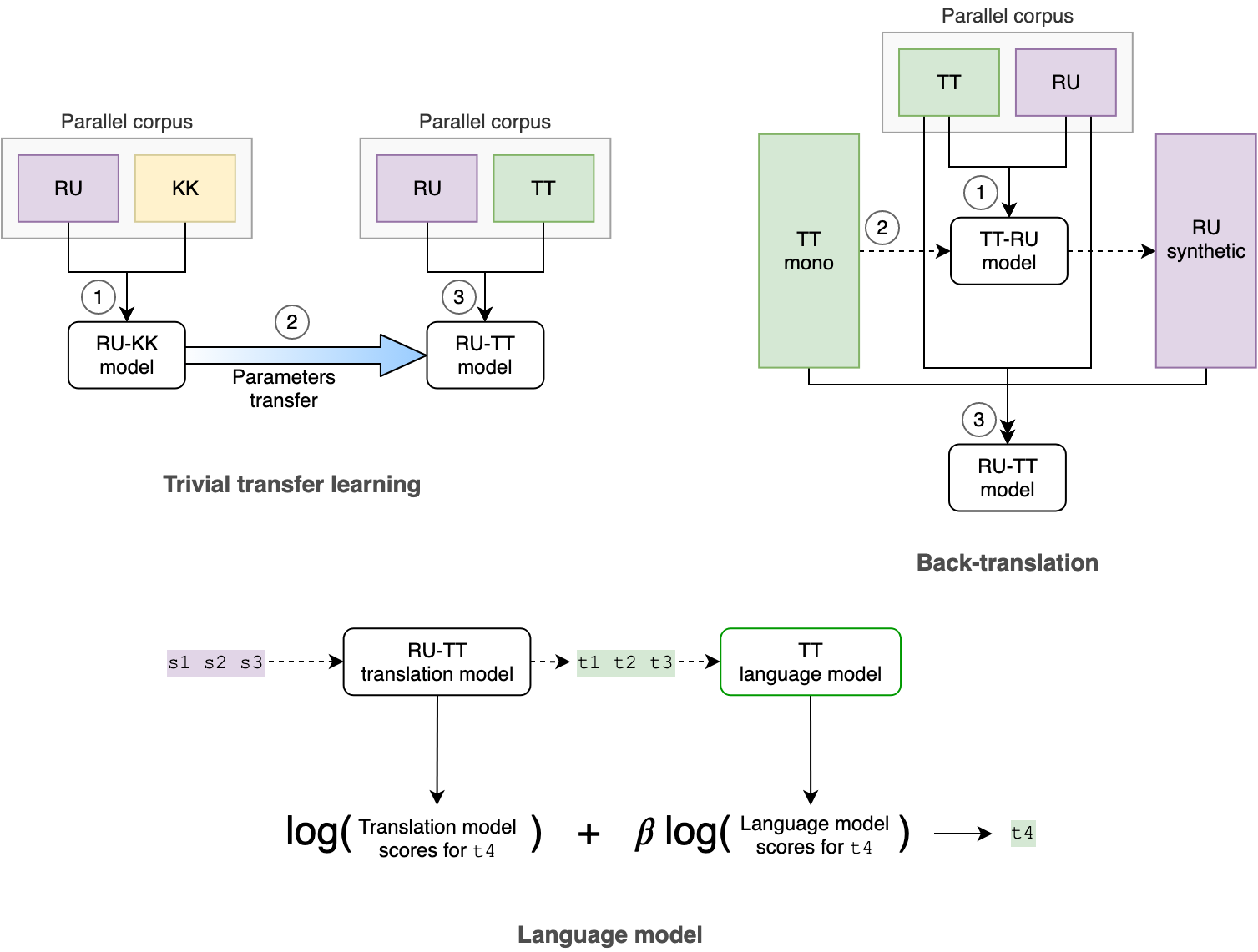}
\caption{The illustration of applied methods. For Trivial transfer learning and Back-translation, the training process is shown. The ultimate goal is to train a Russian-Tatar translation model. In the first case, a parallel corpus with the related Kazakh language is used to obtain parameters for initialization; in the second case, Tatar monolingual corpus is utilized to produce synthetic data. ``Model'' here means a translation model. Thin arrows indicate information used for training, and dashed arrows show the flow of translation when the model is already trained. The numbers in circles denote the order of operations. For Language model, the process of predicting the next target word given an input sequence is presented. Logarithms of probabilities produced by two models are weighted and summed up, and the word with the highest score is selected.}
\label{fig:methods}
\end{figure*}

In this work, we used the approaches described below to improve the translation results for the Russian-Tatar low-resource language pair. For illustrations, refer to Figure \ref{fig:methods}.

\subsection{Trivial Transfer Learning}
To benefit from the related higher-resourced Russian-Kazakh language pair, we apply the approach proposed by Kocmi and Bojar \cite{kocmi2018trivial}. We first train the Transformer model on Russian-Kazakh data, then only change the data to Russian-Tatar and fine-tune. The Byte-Pair Encoding (BPE) \cite{sennrich2016bpe} vocabulary is shared between all three languages. We do this in both directions.

\subsection{Back-translation}
To exploit monolingual data, we apply back-translation as proposed by Sennrich et al. \cite{sennrich2016bt}. We first train a target-source model, then use it to translate the target language monolingual data, mix this synthetic data with the parallel corpus and train the final source-target model on this data. The reverse direction of generating synthetic data is important because this method focuses on improving the target-side language model.

\subsection{Language Model}
To get more out of monolingual data, we implemented Shallow Fusion approach to integrate a language model as proposed by G{\"{u}}l{\c{c}}ehre et al. \cite{gulcehre}. We merged the outputs from the translation model and the language model during beam search controlling the influence of the language model with a hyper-parameter.

\section{\label{section4} Experimental Setup}
\subsection{Data}
We obtained the Tatar-Russian parallel corpus from the Institute of Applied Semiotics, Academy of Sciences of the Republic of Tatarstan, under a non-disclosure agreement. It consists of 324 thousands of parallel sentences. 

As for the Kazakh-Russian parallel corpus, we acquired it from WMT 19\footnote{\url{http://statmt.org/wmt19/translation-task.html}}, and it consists of 5 millions of parallel sentences.

Private Tatar monolingual data was kindly provided by Corpus of Written Tatar\footnote{\url{http://www.corpus.tatar/en}}, and consists of 8.8 millions of sentences.

For Russian monolingual data, we combined news-crawl and news-commentary from WMT 19, resulting in a total of 9 million sentences.

\subsection{Pre-processing}
We deduplicated and divided the parallel data into three parts: validation and testing sets, both of 2K sentences, and training set of 320K sentences. Validation files were used to monitor the convergence of the training. 

BPE with the vocabulary size of 32~768 words was used to produce the vocabulary which is shared between Russian, Tatar and Kazakh languages, if not stated otherwise.

\subsection{System Setup}
We used the Transformer model \cite{vaswani2017transformer} with base parameters from tensorflow github\footnote{\url{https://github.com/tensorflow/models/tree/master/official/transformer}} for the experiments. To evaluate the quality of the translation, we used Bilingual Evaluation Understudy (BLEU) metric \cite{bleu} with max n-gram order equal to 4 and brevity penalty. We will be reporting a case-insensitive percentage BLEU score throughout the paper if it is not mentioned otherwise.

\subsection{Hardware}
Since most of the operations inside the model were numeric and easily parallelizable, NVIDIA GTX 1080 Ti with GPU memory 11 GB was used to speed up the process.

\section{\label{section5} Results and Analysis}
In this section we use TT-RU, RU-TT, KK-RU, RU-KK to refer to Tatar-to-Russian, Russian-to-Tatar, Kazakh-to-Russian, Russian-to-Kazakh translation directions respectively.

\begin{figure*}
\centering
\includegraphics[width=1\linewidth,height=55mm]{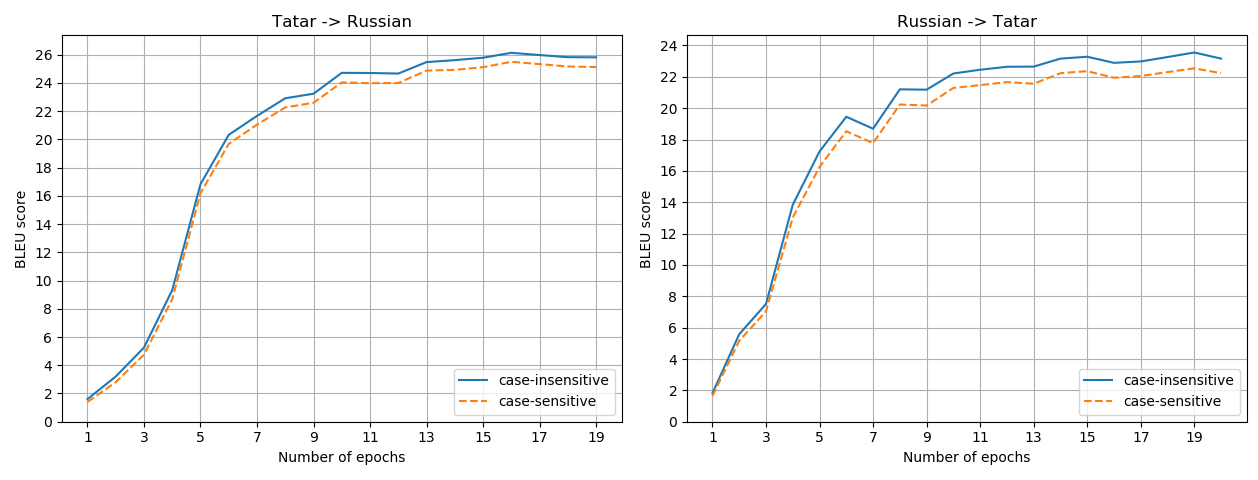}
\caption{Validation scores of the baseline solution}
\label{fig:baseline_bleu_epochs}
\end{figure*}

\subsection{Baseline}
First, we trained the Transformer model on the parallel corpus and got 26.4 BLEU for TT-RU and 23.9 BLEU for RU-TT on the test set; validation scores during training are presented in Figure \ref{fig:baseline_bleu_epochs}. The TT-RU model has a smoother learning curve and reached higher BLUE score. Each epoch took 20 minutes, and we trained TT-RU and RU-TT for 19 and 20 epochs respectively. The vocabulary was only shared between Russian and Tatar.

\subsection{Transfer Learning}
The first approach we tried was transfer learning. The Transformer model trained on the Kazakh-Russian parallel corpus for one epoch gave 53.92 BLEU for KK-RU and 43.83 BLEU for RU-KK, which we considered as very good results, and thus, stopped training. Then we fine-tuned on the Tatar-Russian parallel corpus for 30 epochs getting +2.03 BLEU for TT-RU and +1.82 BLEU for RU-TT. Training on the Russian-Kazakh data takes 5 hours per epoch. In the original paper \cite{kocmi2018trivial}, the method gave different results depending on corpora sizes and languages---our results are good, taking into account that we did not train the model till convergence.

\begin{figure*}
\centering
\includegraphics[width=1\linewidth,height=55mm]{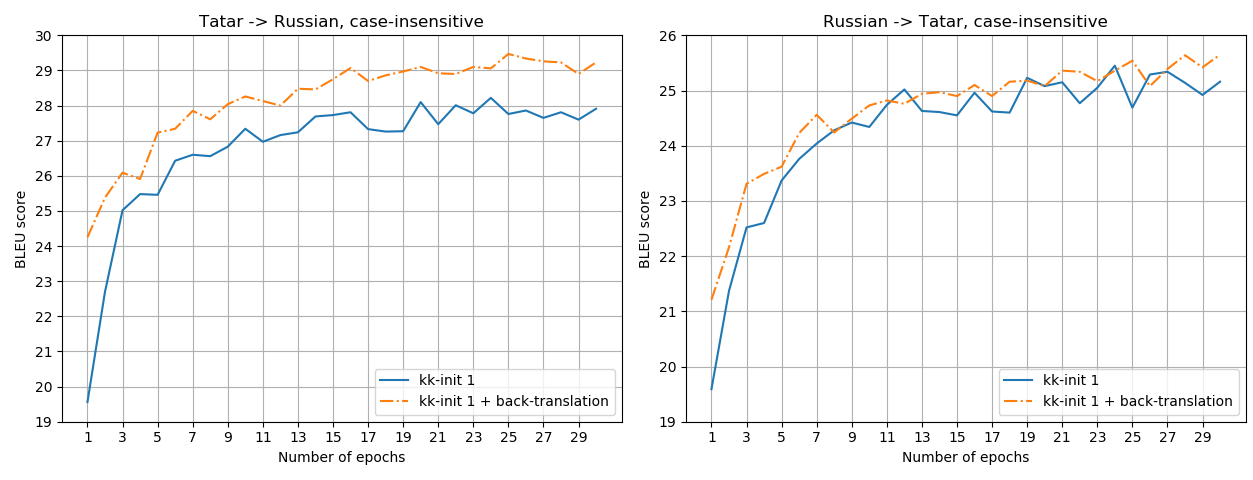}
\caption{Validation scores of transfer learning (Kazakh 1-epoch initialization) and back-translation models.}
\label{fig:bt_kk_init1}
\end{figure*}

\subsection{Back-translation}
Next, we applied back-translation. To do it, we translated a 2.5M sentences subset of Tatar monolingual dataset to Russian using our best model. The translation took about two days. We mixed the synthetic data with the parallel corpus, which resulted in 2.85M sentence pairs, so the ratio of synthetic to parallel is 8:1, as the optimal value found by Stahlberg et al. \cite{stahlberg}. 

We trained the model using this data upon Kazakh initialization for 30 epochs, and each epoch took 2.3 hours. We gained only +0.7 BLEU in the RU-TT direction. In the paper, however, they achieve much better results, of about +3 BLEU improvement. Our low result may be because the quality of the Tatar monolingual data is not good enough or because we used back-translation over transfer learning, since the result of two methods applied together is not guaranteed to be the same as the sum of results when methods are applied separately---the higher is the score, the harder it is to improve it.

Then we did the same for the opposite direction using the best so far RU-TT model---the one with the Kazakh initialization and back-translation. We achieved +1.63 BLEU improvement in the TT-RU direction. The difference with the previous result might be in that we used only transfer learning model to translate Tatar monolingual data, but transfer learning with back-translation model to translate Russian monolingual data; also Russian language always gave better results, when in the target side, it seems that it's easier for the model to learn. 

Validation scores of the models with back-translation compared with the models using only transfer learning are presented in Figure \ref{fig:bt_kk_init1}. We observe that for the TT-RU direction there's a consistent improvement due to back-translation, which is not so for RU-TT. 

\begin{figure*}
\centering
\includegraphics[width=1\linewidth,height=55mm]{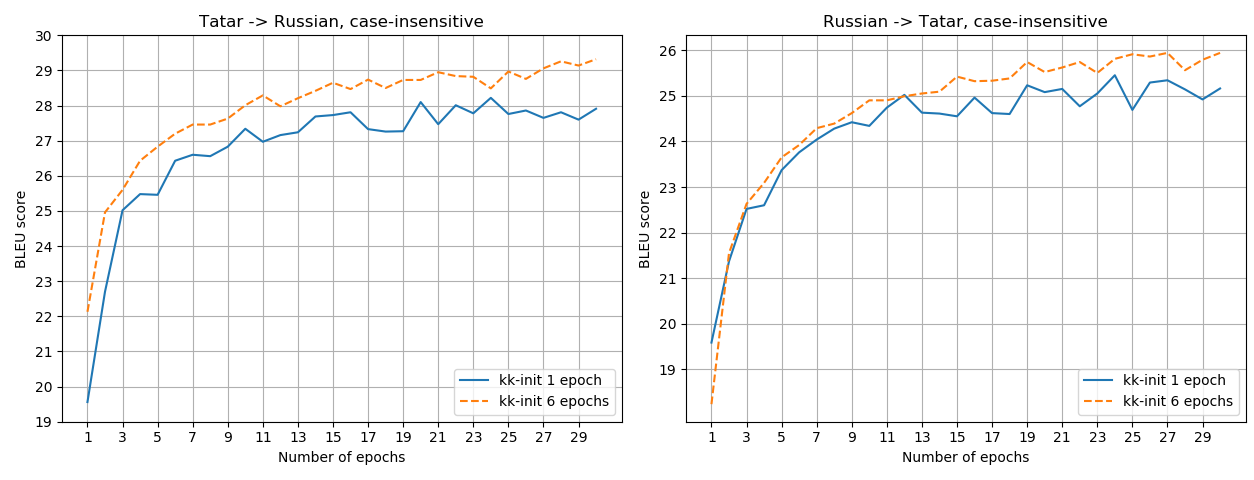}
\caption{Validation scores of transfer learning with Kazakh 1-epoch and Kazakh 6-epoch initialization.}
\label{fig:kk_init6}
\end{figure*}

\subsection{Language Model}
Next, we trained the Tatar language model for Shallow Fusion on the whole Tatar monolingual data for four epochs. We varied the weighting hyper-parameter of the language model when summing its logits with the translation model's logits, but we could not get a significant improvement. With hyper-parameter equal to 0.003, we only obtained +0.004 BLEU (case-sensitive) and +0.01 BLEU (case-insensitive). In the original paper, Shallow Fusion worked differently for different datasets, mostly giving the same insignificant improvements.

\subsection{\label{sectionTLR} Transfer Learning Revisited}

\begin{figure*}
\centering
\includegraphics[width=1\linewidth,height=55mm]{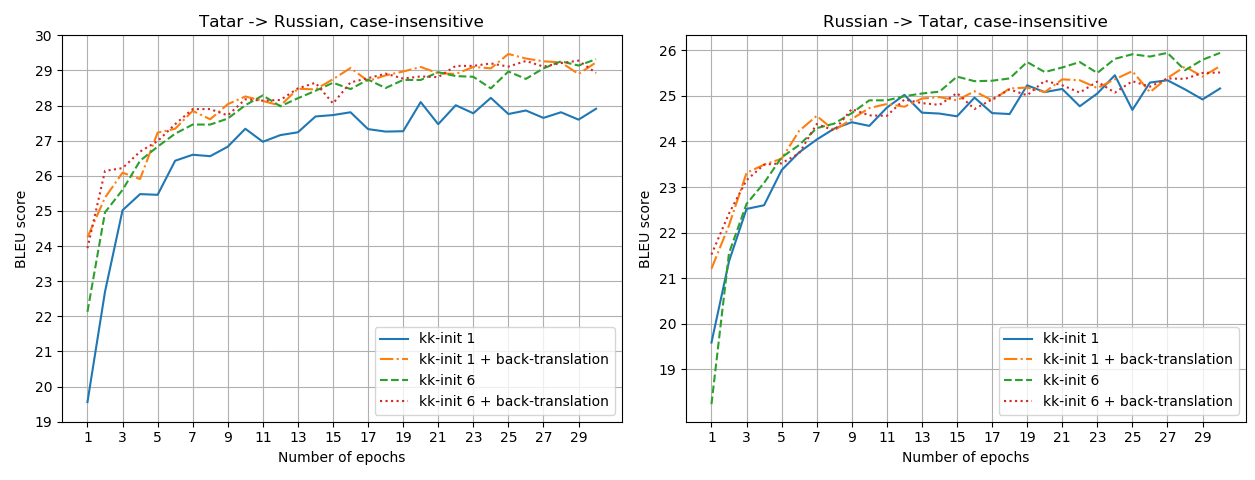}
\caption{Validation scores of transfer learning and back-translation models.}
\label{fig:tr_lr}
\end{figure*}

Now, given that transfer learning yielded the biggest improvements, we wondered whether training on Kazakh-Russian parallel corpus for more than one epoch could improve the results even more. We trained KK-RU and RU-KK models for six epochs achieving 61.14 (+7.22) and 50.24 (+6.41) BLEU, respectively. Each epoch took 5 hours.

Upon this initialization, we trained Tatar-Russian models on parallel corpus for 30 epochs each. Comparing with the same models upon 1-epoch Kazakh initialization, we obtained +0.86 and +0.75 (+2.89 and +2.57 over baseline) BLEU for TT-RU and RU-TT, respectively. Validation scores of the models compared with the 1-epoch initialized one are presented in Figure \ref{fig:kk_init6}.

We also translated the monolingual corpora again using these 6-epochs initialized models and performed back-translation. Interestingly, adding back-translation on top of 6-epochs Kazakh initialization didn't give the expected improvements: for TT-RU it is insignificant (+0.12 BLEU), for RU-TT back-translation even worsened the results (-0.53 BLEU), see Table \ref{tab:test_models_1}. 

We can analyze it deeper by looking at validation scores in Figure \ref{fig:tr_lr}: back-translation doesn't always improve the final result because it doesn't have a simple additive effect. We think it can improve the performance if the underlying model hasn't reached its potential: for example, we can observe that back-translation was very beneficial when it was applied on 1-epoch Kazakh initialization in the TT-RU direction. But when the quality of the model improved (6-epochs initialization), the effect became negligible. For RU-TT, back-translation had some effect on 1-epoch initialized model (better seen in Figure \ref{fig:bt_kk_init1}), but only negatively influenced the better 6-epochs initialized model. It even seems that the quality of the underlying model is inversely proportional to the benefit brought about by back-translation, which may sound counter-intuitive, as we expect better models to generate better synthetic corpus.

Also, the effect of back-translation is highly dependent on the synthetic to parallel data ratio. Varying it, we might achieve better results with back-translation because there could be some dependencies between the optimal ratio and the quality of the model, which was used to translate the monolingual data; the additional experiments are needed to explore them.


\begin{table} 
\footnotesize
\centering
\begin{tabular}{p{2cm}p{1.5cm}cp{1cm}cp{1cm}}
\toprule
\bfseries Model & \bfseries Case & \bfseries TT-RU & \bfseries RU-TT \\
\midrule\multirow{2}{*}{Baseline} & insensitive & 26.40 & 23.90 \\ & sensitive & 25.70 & 22.82 \\
\midrule\multirow{2}{*}{kk-init-1} & insensitive & 28.43 & 25.72 \\ & sensitive & 27.69 & 24.68 \\
\midrule\multirow{2}{*}{bt+kk-init-1} & insensitive & \textbf{30.06} & 26.40 \\ & sensitive & \textbf{29.29} & 25.36 \\
\midrule\multirow{2}{*}{kk-init-6} & insensitive & 29.29 & \textbf{26.47} \\ & sensitive & 28.59 & \textbf{25.45} \\
\midrule\multirow{2}{*}{bt+kk-init-6} & insensitive & 29.41 & 25.94 \\ & sensitive & 28.66 & 24.87 \\
\bottomrule\\
\end{tabular}
\caption{\label{tab:test_models_1} Test scores of all trained models.}
\end{table}

\section{\label{section6} Conclusion}
As demonstrated in Table \ref{tab:test_models_1}, we obtained the most significant improvement due to transfer learning, though we only performed back-translation upon transfer learning, so no fair comparison. We noticed that back-translation can significantly increase translation accuracy, but its performance is not stable, as illustrated in Section \ref{sectionTLR}. From a practical standpoint, translation takes a lot of time, and back-translation increases the training time proportionally to the size of data---8 times longer in our experiments.

We integrated the language model using Shallow Fusion, but we could not get a considerable improvement. This, probably, indicates that a more efficient integration method is needed, e.g., Deep Fusion \cite{gulcehre} or Simple Fusion \cite{stahlberg}.

We observed that depending on the translation direction training goes differently, the techniques we tried also have a different effect. In the end, we obtained the best result for the Tatar-to-Russian direction when we applied back-translation upon 1-epoch Kazakh initialization; for Russian-to-Tatar, 6-epochs Kazakh-initialized model showed the best results. 

To summarize, we explored neural machine translation for the Russian-Tatar language pair and presented our results and analysis of applying Transfer Learning, Back-translation, and Shallow Fusion to the base Transformer model. We experimented with unexplored language pair, at the same time filling the gap in low-resource MT between Slavic and Turkic languages.

\bibliography{my_bib}
\bibliographystyle{IEEEtran}

\end{document}